# Multi Camera Placement via Z-buffer Rendering for the Optimization of the Coverage and the Visual Hull

Maria L. Hänel and Johannes Völkel and Dominik Henrich

*Abstract*— We can only allow human-robot-cooperation in a common work cell if the human integrity is guaranteed. A surveillance system with multiple cameras can detect collisions without contact to the human collaborator. A failure safe system needs to optimally cover the important areas of the robot work cell with safety overlap. We propose an efficient algorithm for optimally placing and orienting the cameras in a 3D CAD model of the work cell. In order to evaluate the quality of the camera constellation in each step, our method simulates the vision system using a z-buffer rendering technique for image acquisition, a voxel space for the overlap and a refined visual hull method for a conservative human reconstruction. The simulation allows to evaluate the quality with respect to the distortion of images and advanced image analysis in the presence of static and dynamic visual obstacles such as tables, racks, walls, robots and people. Our method is ideally suited for maximizing the coverage of multiple cameras or minimizing an error made by the visual hull and can be extended to probabilistic space carving.

## I. INTRODUCTION

Human-robot-cooperation and coexistence is a highly complex and rapidly developing field which addresses matters beyond trajectory planning, such as sensor usage, behavior studies, and safety concerns. During the cooperation with an industrial robot in Figure 1, a human collaborator may be too slow to avoid collisions with the robot. For protection, the environment is often observed by cameras to reconstruct the scene [1], find the collaborator and reschedule the robot's route if necessary [2]. Cameras are preferred as sensors since no tracking device needs to be attached and no direct contact between sensor and human needs to be established.

The usage of *multiple cameras* may yield a better three-dimensional understanding of the environment: The images from each individual camera indicate the location of the collaborator in the environment from one perspective. In order to refine the location, the camera views can be combined in the image space as is typical in stereo vision [3]. More than two sensors are typically combined in the 3D environment [4], for instance as visual hull [5] or scene flow motion estimation [6]. In human-robot collaboration, the visual hull is a suitable choice for integrating multiple camera views because it resembles a conservative reconstruction of a human. Conservative means that the human shape is never underestimated, which is essential to avoid collisions. The conservative visual hull is extended to regard static and dynamic visual obstacles [7] for human-robot-cooperation.

Chair for Robotics and Embedded Systems, University of Bayreuth, 95440 Bayreuth, Germany, maria.l.haenel@gmail.com, dominik.henrich@uni-bayreuth.de

This work has partly been supported by the Deutsche Forschungsgemeinschaft (DFG) under grant agreement He2696/11 SIMERO.

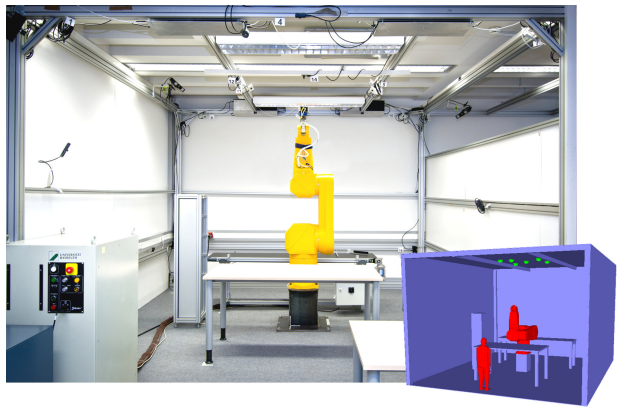

Fig. 1. Left: Robot environment including multiple cameras on the ceiling. Right: An architecture for optimal camera placement is proposed which accepts a 3D CAD-model of the robot cell as input. The basic model has 300 faces, the human model has 85k faces, the robot model 35k faces.

The relevant regions of the environment will be covered more thoroughly and the human reconstruction will be more precise, the better the cameras are placed and oriented. So, in order to reduce false alarms and to deploy a failure resistant system, an optimal camera constellation needs to be found. A virtual reality interface for a manual placement exists [8] but while human consideration yields good result most of the time, cf. Fig. 9, the result may not be optimal.

### A. Aim

Optimal camera placement can be phrased as a non-convex optimization problem: *Given:* A 3D environment (Fig. 1) with human poses, static obstacles such as racks or tables, and dynamic obstacles such as the robot; And given $M \in \mathbb{N}$ cameras. *Domain:* The extrinsic camera parameters (positions and orientations) are the optimization variables. The variables are constrained by the possible mounting area of the environment but are not restrained to discrete mounting spots. *Objective function:* It encodes the quality of a camera constellation. Here, the objective function is a visibility simulation in the presence of static and dynamic obstacles. *Optimization:* A *solver* iterates over the variables to find high quality positions and orientations of the cameras calling the objective in each iteration step in order to

*Max coverage:* Maximize the parts of the environment that are covered by at least $k$ cameras or

*Min error:* Minimize the error of the human collaborator's visual hull

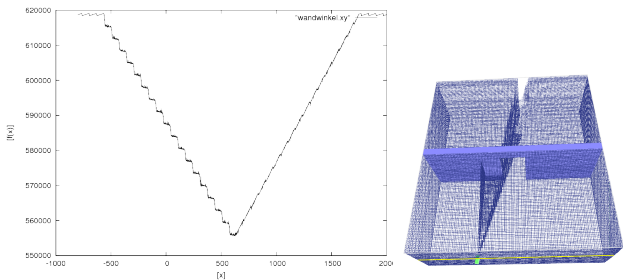

Fig. 2. Example of a quantized, semi-differentiable, and non-convex objective. Right: Setup with a camera (green), 194K voxels (blue outline), and a visual obstacle (a wall with a door, grey). Left: Diagram of the volume of the undetectable space (ordinate) against the horizontal camera position (abscissa, the orientation is constant).

The purpose of this work is to propose a numerical algorithm that automatically solves these two problems.

*B. Content*

Last section showed the general demand for solving the optimal camera placement problem and relevance to Robotics. We will depict related work and demonstrate the originality of our contribution in Sec. II. The Sec. III mathematically outlines the problem and Sec. IV elaborates on the algorithm design. Realistic experiments in Section V show the significance and technical quality of the method.

## II. RELATED WORK

The optimal camera placement methods can be categorized by the type of problem or by their solvers:

*1) General problems:* In case the application is unknown, a general form of optimal camera placement needs to be used. The *Art gallery problem* [9], [10] is geometrically concerned with minimizing the number of cameras such that the environment boundary is observed completely. Its automation has been adressed as an integer problem by minimizing the cameras [11]–[13] or the money [14]. In [15] three methods are compared on this topic, [14], [16], [17]. The reverse problem optimizes the coverage with a fixed number of cameras. The *Maximal covering location problem* [18], [19] and *Facility location problem* [20], [21] are the integer versions. You can optimize the non-linear Max Coverage problem by the uncovered area [22], [23] or vision probability [24] or the trajectory resolution [25], [26] or by the camera-object distance [27].

*2) Application specific problems:* With Robotics in mind, you want to maximize the picture quality [28] or minimize the error caused by a suboptimal multi camera view: There are optimal control methods that dynamically adjust the cameras while performing a (robotic) task, such as eye-in-hand movement [27], [29], [30], scene modelling [22], object reconstruction [31], active stereo tracking [32] or measurement [33], decentralized tracking [34], visual servoing [35], or multi-robot formation [36]. Suitable for motion tracking systems are offline methods that are designed for optimizing and attaching the cameras before executing the task: Some of these do not regard obstacles in the environment [33], [37], [38] some of them do [24], [39]–[41], some of them regard risk maps or uncertainty [25], [42].

The closest publications to this work optimize the visual hull [38]–[40], but [38], [40] regard static obstacles but do not incorporate them in the visual hull and choose cameras out of a predefined set in 2D and [39] maximizes the robot-hull-distance instead of minimizing the residual.

*3) Solvers:* The represented authors use popular optimization techniques including greedy methods [11], [25], [43], [44] or hillclimbing [25] or optimize one camera at a time [21], [25], [27], [45] or a subset of cameras [38]. These methods typically find a local optimum but use derivative information or else stop at a stationary point. Here, the objective function can be quantized, semi-differentiable, and non-convex, as illustrated in Figure 2 which makes it difficult to derive the gradient. Statistical methods applied in [24], [39], [46] do not rely on gradient information, are globally convergent but slow. For fast deterministic global optimality, some authors formulate the optimal camera placement problem as a binary program with different types of visibility and connectivity constraints [47]–[50] and solve it with Branch and Bound [14], [51]. Non-linear constraints such as the visibility of a camera are typically precomputed and provided in matrix formulation, additionally the authors choose camera positions out of predefined sets.

*4) Contributions:* In contrast to recent work, we improve camera vision in 3D applications regarding static and dynamic obstacles. We optimize the poses of multiple cameras on a continuous domain as opposed to integer optimization. We optimize before deploying the cameras in the surveillance system which can be extended to online applications. We do not precompute, we compute the visibility and other constraints during the optimization. Previous work does not simulate the camera pixels which is crucial for integrating distorted images and advanced image analysis in the optimization. Moreover, camera overlap has not been addressed with respect to the visual hull in optimal camera placement.

## III. PROBLEM DEFINITION

Let $E \subset \mathbb{R}^3$ denote the inside of a 3D polyhedral robotic environment. Let $\mathbb{P}$ denote the *parameter space of a single camera*, e.g., $\mathbb{P} = E \times [0, 2\pi] \times [0, 2\pi]$ for three position and two orientation parameters. The visibility simulation in 3D encompasses boolean operations on polyhedra. The geometrical computation of the latter can be non-robust [52], which is why we follow the standard procedure [7], [53] and discretize the environment into smaller cells called *voxels* as in Fig. 2. Let $\mathbb{A} \subset E$ denote the set of all voxels of the environment. The view of a single camera is denoted by the function

$$\sigma_E : \mathbb{P} \to \mathbb{A} \qquad (1)$$

where $\sigma_E(a) \subset \mathbb{A}$ denotes the voxel set visible to the camera with parameters $a \in \mathbb{P}$. The choice of $\sigma$ is explained in Sec IV-B and IV-C.

For $M \in \mathbb{N}$ cameras adjusted with $a_1 \in \mathbb{P}_1, ..., a_M \in \mathbb{P}_M$, we call $x := (a_1, ..., a_M)$ the *variable vector* and $\mathcal{D} := \mathbb{P}_1 \times$

$\ldots \times \mathbb{P}_M$ the *domain* of the optimization problem. Let $1 \leq k \leq M$ denote the *camera overlap*, i.e. the minimum number of cameras by which the important space of the environment needs to be covered. In this work, the objective function $f_k$ is decomposed by (2) into $M$ camera specific functions $\sigma_E$ which merely depend on the parameters of a single camera $a_m \in \mathbb{P}_m$ and each define a voxel set $\mathbb{A}_1, \ldots, \mathbb{A}_M \subset E$.

$$f_k(x) = g_k \circ (\sigma_E, \ldots, \sigma_E)^T (x) \quad (2)$$
$$g_k : \quad (\mathbb{A}_1 \times \ldots \times \mathbb{A}_M) \to \mathbb{R} \quad (3)$$

The metric $g_k$ integrates the views of multiple cameras with respect to the overlap $k$ and evaluates its quality (cf. Sec. IV-D). In this work we consider the problem of the general form

$$\text{Find:} \quad \underset{x \in \mathcal{D}}{\arg\max} \, f_k(x). \quad (4)$$

Depending on the choice of $\sigma_E$, the objective quality $f_k$ of a camera constellation $x$ may resemble (i) the volume of combined camera coverage with overlap $k$ or (ii) the volume of the visual hull error with safety threshold $k$.

## IV. VISIBILITY SIMULATION

To solve either problem, the objective quality $f$ is simulated for each valid variable vector $x$ according to the the application's processing steps. Sec. IV-A is dedicated to localising a target such as the human in the image space of a single camera. With this information Sec. IV-B derives the camera's coverage and Sec. IV-C localises the target in the voxel space of the same camera. The Sec. IV-D integrates the camera views in a combined voxel space, and Sec. IV-E provides the complexity and acceleration.

### A. Image space of a single camera view

*1) Image acquisition::* The images of the robotic real-time application [7], [53] are frames of a video stream (Fig. 3, top). When optimizing the cameras before deploying them, a video stream is not available. For image synthesis, we could assume a pinhole camera model with ray-tracing [39], but distorted images or elaborate image analysis cannot be simulated this way. We suggest rendering an image with $n_x, n_y \in \mathbb{N}$ pixels from the CAD-model of the environment. To accelerate the voxel coloring in Sec. IV-B, we render depth images, where each pixel value corresponds to the distance between the camera position and the next face (Fig. 3, bottom). The image of the camera positioned in $a \in \mathbb{P}$ is defined by an image plane $I(a) \subset \mathbb{R}^3$ and an image boundary $[1, n_x] \times [1, n_y]$. Let $\partial E$ depict the boundary of $E$. To render an image start with an image full of $\infty$ values and follow the z-buffer procedure [54] for all faces of the environment $\partial E$:

*a) First:* project all the vertices $y \in \partial E$ of the face into the image plane $I$ by matrix multiplication $P_a : E \to I$.
*b) If:* one of the projected vertices is in the image $P_a(y) \in [1, n_x] \times [1, n_y]$ then retrieve all their depth values, which is the distance $d(y, a)$. (In case one of the points is outside the image boundary, use the pixels at the boundary).

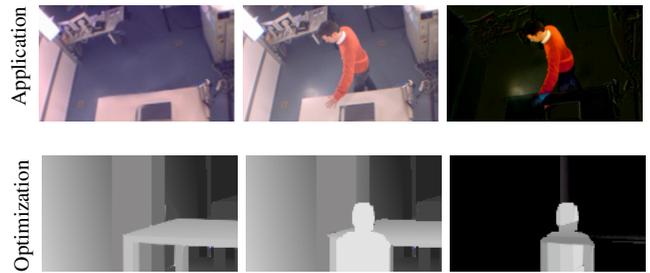

Fig. 3. Top: RGB video frames of the robotic application. Bottom: Synthetic depth images for the camera optimization; Segmentation: An image is taken before (left) and after (middle) humans enter the environment. The value of each pixel in the first image is subtracted from the value of the same pixel in the second image, resulting in a segmented image (right) with *background pixels* (black) and *foreground pixels* (lighter). A person will be segmented as foreground.

*c) If:* one of the depth values is smaller than the already stored values in the pixels, overwrite the pixel and fill the whole face with the interpolated depth values. Here, we assume that the environment has no intersecting faces.
The image is then a function $i_{a,E} : [1, n_x] \times [1, n_y] \to \mathbb{R}$ and if a face $y \in \partial E$ in direction of the pixel $p \in [1, n_x] \times [1, n_y]$ exists, the pixel has the depth value

$$i_{a,E}(p) = \min_{\substack{y \in \partial E \\ p = P_a(y)}} d(y, a) \quad (5)$$

*2) Segmentation::* The human silhouette can be identified in the images of each camera by advanced image analysis methods including Gaussian Mixture Models [1] and regularization. The optimization should include the same technique as the real application. We utilize a *background subtraction* method as illustrated in Figure 3: Let $E_d$ denote the environment with dynamic objects and $E_s$ the environment with only the static objects. The value of the image with static objects in the environment is subtracted from the image including dynamic objects, resulting in the *segmented image s*

$$s(p) = i_{a,E_d}(p) - i_{a,E_s}(p). \quad (6)$$

In applications with real color frames you should distinguish the background pixels by $|s(p)| \approx 0$, but we use rendered depth images, so the following areas are segmented in $s$:

- If the dynamic faces are further away than the static faces $s(p) \geq 0$, the pixel $p$ only shows the static objects which are called the *background*.
- If the dynamic faces are in front of the static faces $s(p) < 0$, the pixel $p$ shows the dynamic faces. These pixels are called the *foreground pixels*. The human target is a dynamic object and is shown as foreground in the segmented image.

### B. Voxel space of one camera for Max Coverage

Here, we prepare the voxel space of a single camera to show its coverage. It can be directly used for Max Coverage skipping Section IV-C or indirectly for Min Error by including it.

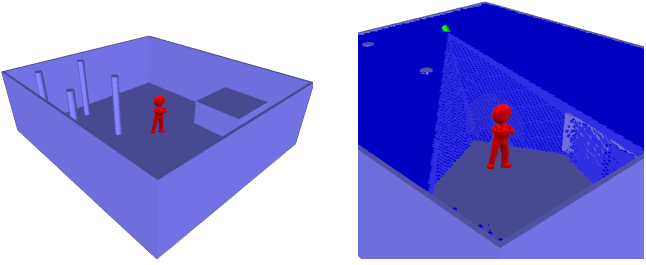
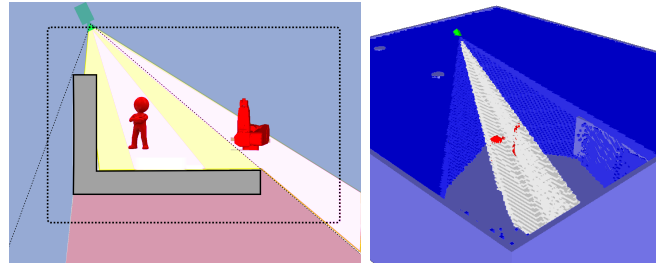

Fig. 4. Voxel coloring for Max Coverage in an environment (left) with static (grey) and dynamic objects (red) by the images of a single camera (right). Given the camera (green) each voxel $y \in \mathbb{A}$ is assigned one of two attributes: The attribute *undetectable* is assigned to voxels outside the camera frustum or behind an object (blue). Else it is detectable (transparent).

Fig. 5. Pictogram (left) and rendered scene with voxel coloring for Min Error in an environment with static (grey) and dynamic objects (red). Given the camera (green) each voxel $y \in \mathbb{A}$ is assigned one of four attributes: *Outside the camera frustum* (blue) or *occluded* behind an object (red). If none of these two, the voxel is *changed* (white) if the segmented image shows a foreground pixel, and *identical* (left: yellow, right: invisible) if it shows a background pixel.

*1) Voxel coloring:* The depth images (5) are used to color the voxel set $\mathbb{A}$, as illustrated in Figure 4. Given the camera with the parameters $a \in \mathbb{P}$, each voxel $y \in \mathbb{A}$ is either assigned the attribute detectable or undetectable:

*a) Undetectable::* We project the voxel $y$ onto the image plane in constant time $\mathcal{O}(1)$ by matrix multiplication $P_a(y)$ and check whether it is outside the opening angle or occluded behind an object:

- *Outside the opening angle:* Check whether the projected voxel is outside the boundary $P_a(y) \notin [1, n_x] \times [1, n_y]$
- *Occluded:* If not we compare the voxel-camera distance $\mathrm{d}(y, a)$ to the depth value $i_{a, E_s}(P_a(y))$ in the image showing all the static objects. If the voxel-camera-distance is larger than the pixel value, $\mathrm{d}(y, a) > i_{a, E_s}(P_a(y))$, the voxel is occluded by static objects.

*b) Detectable:* If the voxel $y$ is not undetectable, it is detectable.

*2) Choice of $\sigma$:* In the end we want to maximize the camera coverage. The voxels that are relevant to the optimization are all voxels in $\mathbb{A}$ that are detectable. So, for the camera perspective defined by $a \in \mathbb{P}$ we choose the set

$$\sigma_{E_s}(a) := \{y \in \mathbb{A} \mid \text{attribute}(y) = \text{'Detectable'}\} \quad (7)$$

Bear in mind that this voxel set depends on the environment. If we used an image with all the objects $i_{a, E_d \cap E_s}(P_a(y))$ instead of the image with the static objects $i_{a, E_s}(P_a(y))$, we would neglect areas behind dynamic objects which can move away after the optimization.

### C. Voxel space of one camera for Min Error

You can skip this section if you want to maximize the coverage of multiple cameras. We prepare the voxel space of a single camera for the conservative visual hull, an approximation of the target that fully encloses it (cf. Sec.I). We consider the target to be a dynamic visual obstacle. As an intuitive explanation, we identify the space of the environment in the images that cannot contain the target. With each camera perspective we carve out more space of the environment to refine the location of the target.

*a) Voxel coloring::* The areas with different voxel color are illustrated in Fig. 5. Given the camera with the parameters $a \in \mathbb{P}$, the voxel $y \in \mathbb{A}$ is again projected into the image plane $P_a(y)$.

*Undetectable:* (blue and red) The static depth images (5) are used as in Sec. IV-B to distinguish between detectable and undetectable voxels.

*Detectable:* If the voxel $y$ is detectable, we assign the following attributes according to the segmented image(6) including both static and dynamic faces:

- *Identical:* (yellow) Check if the projected voxel is in a background area $s(P_a(y)) \geq 0$. Reversely, each background area in the segmented image produces a ray of identical voxels from the camera to the next static face of the environment.
- *Changed:* (white) Check if the projected voxel is foreground in the segmented image $s(P_a(y)) < 0$.

*b) Choice of $\sigma$::* The system has detected dynamical objects in the 'changed' space and it does not have enough information whether the 'undetectable' space includes dynamical objects. However, it can testify that the 'identical' space has not changed and is dynamical-object-free. Thus a conservative visual hull of the dynamical objects includes all the 'changed' and 'undetectable' voxels.

$$\sigma_{E_d}(a) := \{y \in \mathbb{A} \mid \text{attrib}(y) = \text{'Changed'} \text{ or } \text{'Undet.'}\}. \quad (8)$$

Bear in mind that this voxel set depends on the environment. In a given environment, the set only depends on the parameters of the camera $\sigma_{E_d}(a) \equiv \sigma(a)$.

### D. Voxel space of multi camera views

*a) Combined view for Max Coverage:* In order to maximize the coverage, the voxel sets of $M$ camera views $\sigma(a_m)$, $m = 1, \ldots, M$ are integrated. The intersection of the 'detectable' voxels (7) in $\mathbb{A}$, $\sigma(a_1) \cap \ldots \cap \sigma(a_M)$ means all the cameras need to cover the space of the environment, the union of these voxels means any of the cameras is allowed to cover it. The overlap threshold $k \in \mathbb{N}$ with $k \leq M$ (Sec. III) makes sure that only voxels are included that are included in at least $k$ camera views. The *multi camera view* is the voxel set covered by at least $k$ cameras and is defined by

$$\mathcal{C}_k(x) = \mathcal{C}_k(a_1, ..., a_M) := \bigcup_{\pi \in c_M^k} \bigcap_{m=1}^{k} \sigma(a_{\pi(m)}) \quad (9)$$

with $c_M^k \subset S_M$ denoting the set of $k$-combinations of the $M$-tuple $(1, ..., M)$ and the order $\frac{M!}{k!(M-k)!}$ (as opposed to $M$-permutations $S_M$).

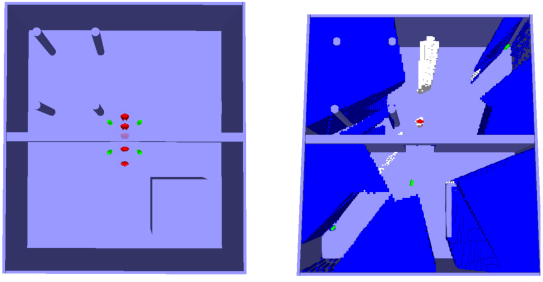
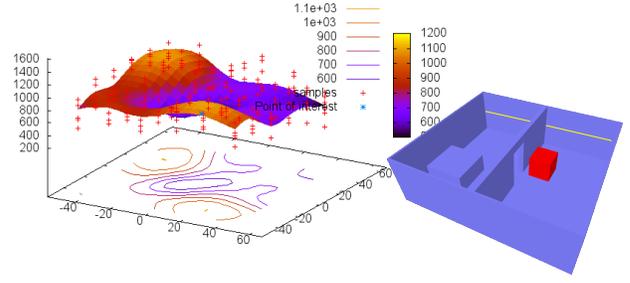

Fig. 6. Voxel coloring in an environment with static (grey) and dynamic objects (red) in $t = 1, ..., 5$ time steps, $M = 4$ cameras. Each voxel $y \in \mathbb{A}$ is assigned one of three attributes: *Undetectable by all the cameras* (blue) or *identical* (transparent) if it can be identified as background in at least $k = 1$ camera views or *changed* (white) if it is foreground or undetectable in $M - (k - 1) = 4$ camera images.

Fig. 7. Right: Test environment, where $M = 3$ cameras are placed in six steps along the yellow line aimed perpendicular to the wall: $x \in [-45, 45]^3$. Left surface: Non-convex but differentiable RBF-interpolant of the volume of the multi camera view (ordinate) with respect to the positions of two cameras (abscissa). The volume of the third camera can be seen as crosses.

*b) Metric::* The metric $g_k : (\mathbb{A}_1 \times ... \times \mathbb{A}_M) \to \mathbb{R}$ evaluates the quality of the multi camera view. For instance,

$$g_k(x) = \#\mathcal{C}_k(x) \quad (10)$$

counts the voxels in the multi camera view. In case, some areas are more important, you can easily assign a scalar field weighing the voxels similar to applying risk maps [42]. Moreover, multiplied by a specific constant you can easily derive the volume of the voxel space.

*c) Combined view for Min Error:* In order to form the visual hull, the 'Changed' or 'Undetectable' voxel sets (8) of all the cameras are integrated in the same way as in (9) above, i.e. the multi camera view is the visual hull. Since the visual hull is constructed as a conservative approximation of the target (i.e. the target is the minimum boundary), minimizing the volume of the multi camera view (9) means minimizing the error of the visual hull.

In case of formulating the problem as a maximization (4) instead of minimization, we need a transformation to the voxels in $\mathbb{A}$ that are definitely not occupied by dynamical objects, i.e. 'Identical' voxels. If the multi camera view $\mathcal{C}_k^{\text{color}}$ integrates at least $k$ camera views $\sigma^{\text{color}}$ with the attribute 'color', one can prove that it holds all the voxels that are not in the multi camera view $\mathcal{C}_{M-(k-1)}^{\text{not color}}$ assigned with any attribute but the color at least $M - (k - 1)$ times

$$\mathcal{C}_k^{\text{color}}(x) = \mathbb{A} \setminus \mathcal{C}_{M-(k-1)}^{\text{not color}}(x). \quad (11)$$

For instance, in Fig. 6 the multi camera view $\mathcal{C}_k(x)$ denotes the set of voxels which are 'identical' in at least $k$ camera views, the rest of the voxels is not 'identical' in at least $(M - (k - 1))$ camera views.

*E. Complexity*

We assume that the CAD-model of the scene is bounded by $f \in \mathbb{N}$ faces sorted in a tree-structure. Let the set of all voxels $\mathbb{A}$ be composed of $v \in \mathbb{N}$ voxels and the images of $p \in \mathbb{N}$ pixels. A depth image of the faces can be produced in $\mathcal{O}\left(p \cdot \log(f)\right)$ [54]. The background subtraction segmentation has a complexity of $\mathcal{O}(p)$ and the coloring and the integration step both have $\mathcal{O}(v)$. The complexity of the complete visibility analysis with $M$ cameras is

$$\mathcal{O}\Big(M \cdot \big(p \cdot \log(f) + p + v\big)\Big) \quad (12)$$

To to execute a single function call $f(x)$, the costs need to be expended for each discretized time step of the dynamical objects.

*a) Additional Acceleration:* The simulation of $\sigma(\cdot)$ by z-buffer rendering is less expensive than raytracing if you intend to color the voxels with it. Additionally, the image rendering and segmentation can be parallelized (in the pixels) by modern graphics cards which makes this approach ultimately valuable. To decrease the average asymptotic behavior even further, one can voxelize the objects beforehand [55] or color voxels first with a higher impact on the measure $g$. The construction of the multi camera view over several time steps is accelerated by regarding spatial and temporal coherency [2] and by stopping the coloring according to the overlap threshold $k$.

V. EXPERIMENTS

We include test results of the optimization of the problem (Sec. III) with the simulation (Sec. IV) in this section. We have optimized the Max Coverage with three cameras on a differentiable vision function in Section V-A. The realistic quantised examples with five cameras are solved in Section V-B for minimizing the error of the visual hull.

*A. Differentiable objective*

*1) Experiment:* the quality function (10) with three cameras and a position variable each $x = (a_1, a_2, a_3)$ was simulated in $6^3$ points of the domain before the optimization (Fig. 7). The function with multiple local and global optima serves as ground truth and allows us to measure how often a solver reached the global optimum. Moreover, the differentiable interpolated function allows us to compare a statistical solver (Ant Colony Algorithm, [57] and [39]) and the surrogate method (CorsRBF, [58] and Appendix) that both do not use derivative information with a deterministic method (SQP [56]) that uses derivative information.

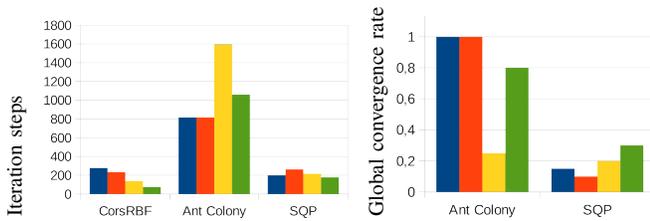

Fig. 8. Left: Bar plot displaying the average number of function evaluations (ordinate) of three solvers (abscissa), the smaller the number the faster the solver. Colors indicate the initial variable vectors: The eight domain boundary vertices (I, blue), the four vertices of the simplex $S$ (II, red), eight points forming a cube inside $S$ (III, yellow), and 20 points uniformly distributed in $S$ (IV, green).
Right: Bar plot of the ratio of tests (ordinate) in the same groups that reached the global maximum. The higher the number, the more tests have reached the global optimum. The CorsRBF has always reached the global optimum and is not shown.

The solvers were tested in four groups (I – IV) varying in the choice of the starting point: (I) Eight corners of the domain $\{\pm 45\} \times \{\pm 45\} \times \{\pm 45\}$. (II) Four vertices of the simplex $S = \{-45 \leq a_1 \leq a_2 \leq a_3 \leq 45\}$ (III) Eight points forming a cube in $S$: $\{-30 \pm 15\} \times \{\pm 15\} \times \{30 \pm 15\}$ (IV) 64 points uniformly distributed in $S$: $\{\pm 30 \pm 15\} \times \{\pm 30 \pm 15\} \times \{\pm 30 \pm 15\}$ While the CorseRBF was started with all the initial variable vectors of one group at the same time (it needs to be initialized with at least $(n + 1 = 4)$ variable vectors), the other solvers were tested with each initial point consecutively. The SQP and ant colony results were averaged over all test runs not exceeding the maximum number of steps (10000).

*2) Result:* The test results are displayed in Fig. 8. In our work, the Ant Colony Solver evaluated the objective more than three times as often as the CorsRBF and SQP. In [39] the same solver used up to 100 times of the number of steps. However, while the CorsRBF always and the Ant Colony Solver most of the time found the optimal solution (termination value is 1144.9), the SQP found the optimum in about one in five test runs. The test also shows a dependency on the choice of the initial variable vectors: While the number of SQP steps are indirectly proportional to the number of initial variable vectors, the CorsRBF and Ant Colony prefer the four ordered vertices of the simplex (red) over the eight vertices of the domain (yellow).

### B. Realistic quantized objective

*1) Experiment:* The visual hull improves with decreasing the number of 'changed' and 'undetectable' voxels. Five cameras were optimally placed regarding the overlap $k = 5$. Each solver step, the simulation (9) is called in a scene with about 600 static object faces and 35k dynamic object faces. The environment is rendered in images of the size $320 \times 240$ pixels. The voxel grid $\mathbb{A}$ has a resolution of $120 \times 135 \times 40$ voxels. Two variables of each camera determine its position on the ceiling $[-35, 35] \times [-40, 40] \times \{26.5\}$. The cameras were directed to a point $(0, 0, 10)$. The initial positions of the cameras for our method were $(\pm 5, \pm 5, 26.5)$. We compared methods not using derivative information, i.e. a hill climbing method (LN_NELDERMEAD [59]) a second order method (LN_NEWUOA [60]), a global surrogate method (CorsRBF, Appendix) and a manual placement: Four cameras in the corners of the room and one directly above the dynamic object. A uniform placement has also been used in [25], [37], [38], [40] for comparison.

*2) Result:* The progress of the solvers is depicted in Fig. 9. As in [25], [37], [38], [40], the uniform placement is good but can be improved. The use of a surrogate already pays off after the 12th iteration step, since a set of starting points can be defined that indicate the improvement direction. Moreover, previous objective values are cached and intermediate values which are not previously evaluated can be approximated without executing $f$. The local methods take longer to overtake the manual placement (around the 60th iteration step). In hindsight the heuristic we chose can be a good initialization vector to accelerate the convergence. Only a few publications include the run time: The next best view in 120 min [22], 16 min to 27 h in total [15], or 6 min to 13 h in total [19]. In this work, no optimization exceeded 24 min in total.

## VI. CONCLUSION

We provide an efficient method for optimally placing multiple cameras in a detailed 3D CAD model. The manual placement is a good initial setting but in the end all the solvers outperformed it. The visibility simulation is ideal for distorted images, an elaborate image analysis, and a voxel grid for the transfer of image data into the real world. Z-buffer rendering accelerates the image acquisition and rendering depth images accelerates the voxel coloring. The surrogate solver exceeds previous methods in optimal camera placement since it has a low number of iteration steps and converges globally. We can optimize any combination of the coverage, the camera overlap and the visual hull.

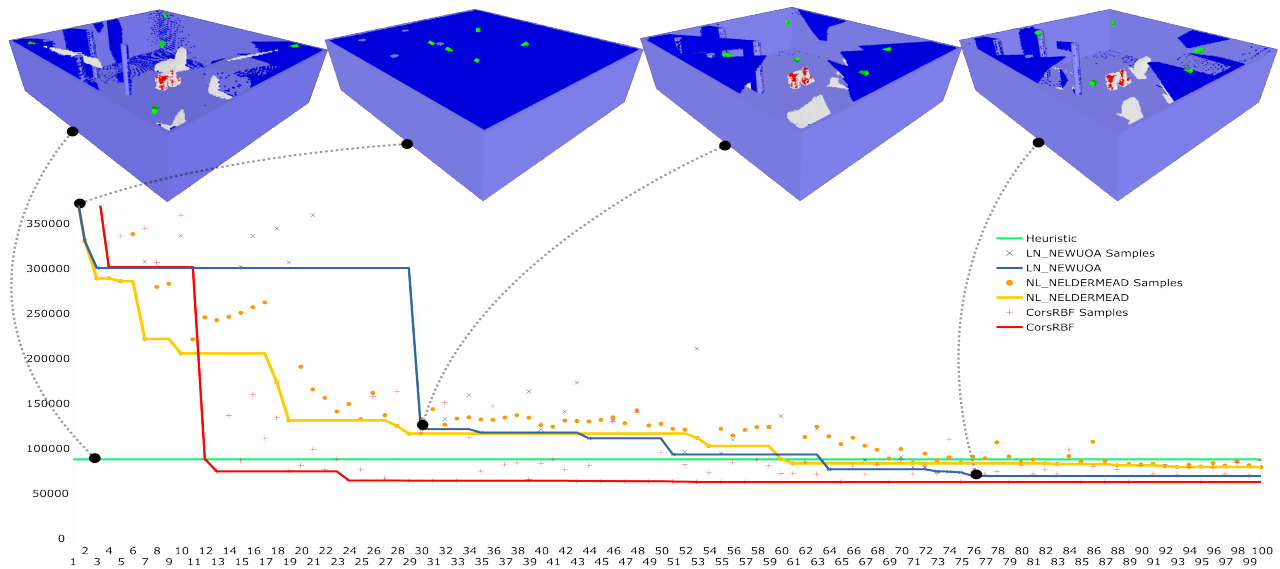

Fig. 9. Bottom diagram: Progress of the error of the visual hull (ordinate) when optimally placing five cameras (three variables each) with a surrogate (CorsRBF, Appendix, red), a quadratic approximation (LN_NEWUOA of [56], blue) or the downhill-simplex (LN_NELDERMEAD of [56], yellow). The smaller the objective value, the more accurate is the visual hull.
Top: Environments displaying a uniform camera placement (left) and the solver progress (second-fourth). Each environment encompasses a static CAD model (grey), a dynamic target (red) and $M = 5$ cameras (green). We can see the 'undetectable' space of all the cameras (overlap $k = 5$, blue) and the conservative visual hull (overlap $k = 5$, white). The detectable target-free space according to at least one camera $k = 1$ is transparent. Potentially, the target can be in all the displayed space (white or blue) which is the reason why white AND blue is minimized.

## APPENDIX

In Optimization, a *surrogate of the objective function* is a differentiable function $\bar{f}$ close to the original possibly quantised, semi-smooth, expensive objective $f$ whose function value and derivative are cheaply evaluated. We implemented a surrogate method [58] consisting of three steps:

(i) Searching for a high quality variables $x$ on the surrogate $\bar{f}$ away from previously evaluated points.
(ii) As soon as a high quality camera constellation is found, the original objective $f(x)$ is evaluated.
(iii) The surrogate is updated so that it interpolates all the previous and the new sample pair $(x, f(x))$.

Our simulation of Sec. IV is only used in step (ii). Other than that, we used the following parameters and libraries for the implementation: The surrogate is the radial basis function (RBF) which interpolates scattered sample pairs [58]. For the update of the surrogate (iii) a linear equation system of size $K + n + 1$ needs to be solved with $n = \dim(\mathcal{D})$ being the size of the problem and the number of previous sample pairs $K$. The LU-decomposition of the gsl allows both, calculating the determinant for exception handling and solving the equation system. Some LU-decompositions are even known to reduce the usual cubic complexity of adding single lines and columns to the system [61].

The optimization step (i) has non-linear constraints that need to be modified in each CorsRBF-iteration step, which is done with IPOPT, in this work. In order to control the size of the areas which are excluded from the domain during the search (i), the furthest distance $\Delta \leftarrow \max_{x \in \mathcal{D}} \min_{1 \leq k \leq K} ||x - s_k||$ is calculated. We used an MMA implemented by nlopt to get this distance. To pick an initial point for the MMA where the objective is differentiable, a point is randomly chosen in the domain excluding the previous iterates. The size of the exclusion areas is then varied relatively to the domain in cycles of five iterations with the ratios $< 0.98, 0.6, 0.75, 0.2, 0.01 >$.